\documentclass[runningheads]{llncs}
\usepackage[T1]{fontenc}
\usepackage{amsmath} 
\usepackage[utf8]{inputenc}
\usepackage{graphicx}
\usepackage{booktabs} 
\usepackage{hhline}
\usepackage{multirow}

\usepackage{marvosym} 
\usepackage{amssymb}

\begin{document}
\title{Audio-Guided Dynamic Modality Fusion with Stereo-Aware Attention for Audio-Visual Navigation}
\titlerunning{AGDF with SAM for Audio-Visual Navigation}
%
\author{
  \small Jia Li\inst{1} \and
   \small  Yinfeng Yu\inst{1}\textsuperscript{(\Letter)} \and
    \small Liejun Wang\inst{1} \and
    \small Fuchun Sun\inst{2} \and
    \small Wendong Zheng\inst{3}
}

\authorrunning{JL, YYF et al.}
%
%

\institute{
   \small  Xinjiang Multimodal Intelligent Processing and Information Security Engineering Technology Research Center, \\
    \small School of Computer Science and Technology, Xinjiang University, Urumqi 830017, China \\
  \email{yuyinfeng@xju.edu.cn} \and
   \small  Department of Computer Science and Technology, Tsinghua University, Beijing 100091, China \and
   \small  School of Electrical Engineering and Automation, Tianjin University of Technology, Tianjin 300382, China
}

\renewcommand{\thefootnote}{}  
\footnotetext{\textsuperscript{(\Letter)} \small Yinfeng Yu is the corresponding author (e-mail: yuyinfeng@xju.edu.cn).}

\maketitle              
\begin{abstract}

In audio-visual navigation (AVN) tasks, an embodied agent must autonomously localize a sound source in unknown and complex 3D environments based on audio-visual signals. Existing methods often rely on static modality fusion strategies and neglect the spatial cues embedded in stereo audio, leading to performance degradation in cluttered or occluded scenes. To address these issues, we propose an end-to-end reinforcement learning-based AVN framework with two key innovations: (1) a \textbf{S}tereo-Aware \textbf{A}ttention \textbf{M}odule (\textbf{SAM}), which learns and exploits the spatial disparity between left and right audio channels to enhance directional sound perception; and (2) an \textbf{A}udio-\textbf{G}uided \textbf{D}ynamic \textbf{F}usion Module (\textbf{AGDF}), which dynamically adjusts the fusion ratio between visual and auditory features based on audio cues, thereby improving robustness to environmental changes. Extensive experiments are conducted on two realistic 3D scene datasets, Replica and Matterport3D, demonstrating that our method significantly outperforms existing approaches in terms of navigation success rate and path efficiency. Notably, our model achieves over 40\% improvement under audio-only conditions compared to the best-performing baselines. These results highlight the importance of explicitly modeling spatial cues from stereo channels and performing deep multi-modal fusion for robust and efficient audio-visual navigation.

\keywords{Multi-modal Fusion \and Embodied AI \and Audio-Visual Navigation.}
\end{abstract}
\section{Introduction}

With the rapid development of artificial intelligence and robotics~\cite{durante2024agent,iftikhar2024artificial,jaquier2025transfer,li2022remote}, the problem of autonomous navigation for embodied agents in virtual environments has gained increasing attention~\cite{chen2023omnidirectional,shi2025towards,wen2025zero,wu2024embodied,yu2023echo,zhao2025audio}. Audio-Visual Navigation (AVN), as a representative multimodal perception and decision-making task, requires an agent to locate a target using auditory cues alone, without access to explicit coordinate information\cite{debner2025towards,wu2024vision,yu2023measuring}. This task poses significant challenges in multimodal information integration and cooperative reasoning.

From a bio-inspired perspective, humans often rely on both visual and auditory cues for spatial understanding during navigation~\cite{wei2022learning,yu2025dgfnet}: visual inputs offer structural and static geometric information about the environment, while auditory inputs are advantageous for directional perception, occlusion handling, and target responsiveness. Thus, effective fusion of these two modalities is critical to improving navigation performance. However, existing methods still exhibit two significant limitations. First, many approaches~\cite{chen2020soundspaces,chen2021waypoints} ignore the spatial disparity inherent in binaural audio by treating stereo signals as undifferentiated input to policy networks, limiting the agent’s ability to infer sound direction. Second, most fusion strategies are static or pre-defined, lacking the adaptability to dynamically adjust modality importance in changing environments, thereby constraining robustness and generalization.

In realistic scenarios, sound sources may be occluded (e.g. a phone ringing behind a closed door) or in reverberant spaces (e.g. hallways, atriums), where visual information cannot directly reveal the source location. In such cases, the agent must rely on subtle spatial differences between audio channels for directional inference. Moreover, the relative importance of vision and audition can vary across different environments: visual cues may dominate in open spaces, while auditory cues are essential in cluttered or low-light conditions~\cite{zellers2022merlot}. Hence, a navigation model must possess strong spatial reasoning capabilities and the ability to adaptively fuse multimodal features according to environmental changes to achieve robust and efficient navigation.

To this end, we propose the \textbf{A}udio-\textbf{G}uided Dynamic Modality Fusion with \textbf{S}tereo-\textbf{A}ware Attention (\textbf{AGSA}) framework, which integrates a Stereo-Aware Attention Module (SAM) and an Audio-Guided Dynamic Fusion Module (AGDF) to enhance the agent's spatial awareness and adaptability. Specifically, the SAM explicitly models the spatial relationship between left and proper audio channels via channel-wise feature separation and bi-directional cross-attention, improving sound source localization. The AGDF dynamically modulates the fusion of visual and auditory features guided by audio semantics and introduces a gating mechanism for adaptive modality interaction. Comprehensive evaluations on the Replica~\cite{replica19arxiv} and Matterport3D~\cite{Matterport3D} datasets show that our method achieves significant gains across standard navigation metrics (e.g. SPL, SR), with over 40\% improvement under audio-only settings compared to competitive baselines. Furthermore, the proposed framework generalizes well to unseen environments, validating its robustness and practical potential.

The main contributions of this work are summarized as follows:
\begin{enumerate}
    \item A Stereo-Aware Attention Module (SAM) is proposed to explicitly model the spatial dependency between binaural audio channels, enhancing directional sound perception in AVN.
    \item An Audio-Guided Dynamic Fusion Module (AGDF) is designed to adaptively integrate audio-visual features, enabling dynamic fusion based on auditory guidance.
    \item Extensive experiments on Replica and Matterport3D demonstrate the effectiveness and generalization capability of the proposed AGSA framework, with significant improvements over existing methods.
\end{enumerate}

\section{Related Work}

\subsection{Audio-Visual Navigation}

Audio-Visual Navigation (AVN) is an embodied AI task that requires an agent to infer and plan paths in real time based on perceived auditory signals and visual cues, ultimately reaching the sound-emitting target. To support this task, several high-quality simulation platforms have been developed. VAR~\cite{dean2020see} builds an indoor audio-visual navigation environment based on AI2-THOR~\cite{kolve2017ai2}, while SoundSpaces~\cite{chen2020soundspaces} leverages Habitat~\cite{savva2019habitat,szot2021habitat} to create a 3D simulation environment with realistic optical and acoustic rendering. SoundSpaces2.0~\cite{chen2022soundspaces} further supports continuous audio-visual perception in arbitrary 3D mesh scenes.

On the methodological front, many works adopt reinforcement learning (RL) to train agents for multi-modal decision making. For example, AV-Nav~\cite{chen2020soundspaces} uses PPO~\cite{schulman2017proximal} to train an LSTM-based policy network that takes both RGB/depth images and audio spectrograms as input. SAVi~\cite{chen2021semantic} incorporates a sound direction prediction module and a temporal Transformer to handle transient audio goals. Other works integrate map-based reasoning or waypoint prediction, such as~\cite{chen2021waypoints,dean2020see}, which leverage occupancy maps and graph search for improved long-range navigation. And~\cite{chen2023omnidirectional} distills point-goal planning into the audio-goal agent for better performance under few-shot settings. In complex audio environments, SAAVN~\cite{YinfengICLR2022saavn} introduces adversarial audio distractors to enhance robustness, while CMHM~\cite{younes2023catch} explicitly models the spatial geometry from both modalities to track moving sound sources under acoustic uncertainty.

Despite the promising progress, most existing approaches still face limitations in spatial awareness and effective modality fusion. The proposed AGSA addresses these gaps by enhancing directional sound localization and adaptively balancing visual-auditory contributions.

\subsection{Feature Fusion}
In AVN tasks, vision and audition provide complementary cues: visual signals reveal environmental structures, while audio cues indicate the direction of the sound source. Effectively fusing both modalities is crucial for navigation performance\cite{gandhi2023multimodal}. Existing works often adopt simple fusion strategies such as concatenation, weighted averaging, or basic attention mechanisms. For instance, AV-Nav~\cite{chen2020soundspaces} encodes RGB/depth images and audio spectrograms independently and concatenates them as policy input. Similarly, many SoundSpaces-based methods~\cite{chen2021semantic,chen2021waypoints,chen2023omnidirectional,roman2025generating,wang2023learning,YinfengICLR2022saavn} use static fusion structures for modality integration. However, such static strategies struggle in dynamic or occluded environments, where modality bias may lead to navigation failure.

Recently, a few works have introduced semantic-aware or context-aware fusion mechanisms to enhance multimodal representation. SAVi~\cite{chen2021semantic} employs a Transformer to model temporal context for better decision-making. FSAAVN~\cite{Yu_2022_BMVC} proposes a context-aware audio-visual fusion strategy via self-attention to track moving targets, focusing explicitly on fusion dynamics in AVN. Nevertheless, these methods require heavy computation and large datasets for stable training.

To address these limitations, we propose an AGDF module. It uses the current audio features as guidance to dynamically modulate the contribution of vision and audition through a gating mechanism. This allows the navigation policy to adaptively adjust modality weights under changing environments, improving robustness and generalization.

\begin{figure}[htbp]
\centering
\includegraphics[scale=0.5]{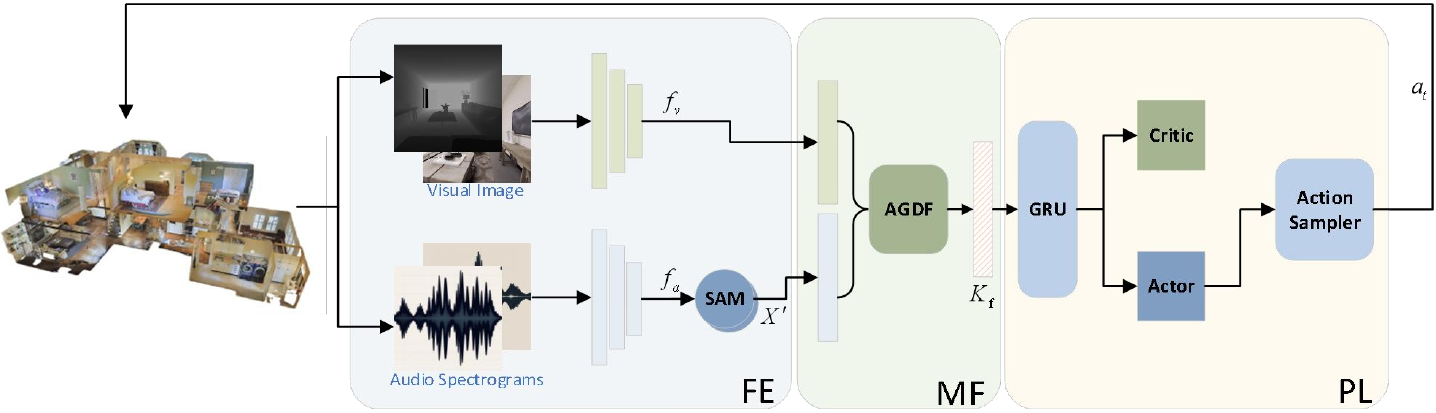}
\caption{The overview of Audio-Guided Dynamic Modality Fusion with Stereo-Aware Attention for Audio-Visual Navigation, where FE: Feature Extraction; MF: Modal Fusion; PL: Policy Learning.}
\label{figure1}
\end{figure}

\section{Method}
\subsection{Overall Framework}

Our method is built upon the standard AudioGoal~\cite{chen2020soundspaces} setting, where an agent must navigate to a continuously sounding source in a complex 3D environment without map information, relying solely on audio and visual perception~\cite{chen2021semantic,wang2023learning}. To improve navigation performance and generalization, we propose the \textbf{Audio-Guided Dynamic Modality Fusion with Stereo-Aware Encoding (AGSA)} framework, which addresses the problem from two perspectives: spatial audio encoding and modality fusion. As shown in Fig.~\ref{figure1}, the framework consists of the following three components:

\begin{enumerate}

    \item [$\bullet$] \textbf{Feature Extraction (FE):} Visual (RGB or depth) and auditory (binaural audio spectrograms) inputs are processed separately by two CNN encoders with identical architecture but independent parameters to extract high-level semantic features. The audio encoder incorporates a \textbf{Stereo-Aware Attention Module (SAM)} to explicitly model spatial correlations between the left and right audio channels, enhancing the agent’s ability to perceive sound directions.
    
    \item [$\bullet$] \textbf{Modality Fusion (MF):} We design an \textbf{Audio-Guided Dynamic Fusion Module (AGDF)} that uses audio features as queries to guide the extraction of supplementary semantic information from the concatenated audio-visual features. A gating mechanism then adaptively fuses these features, enabling flexible and dynamic modality weighting.
    
    \item [$\bullet$] \textbf{Policy Learning (PL):} The fused features are input to a GRU-based temporal encoder and then passed to an Actor-Critic policy network optimized via the PPO algorithm, learning effective navigation behaviors.
    
\end{enumerate}

The agent repeatedly executes the perception–fusion–decision loop until it reaches the sounding source or exceeds the maximum step limit (500 steps). We now describe each component in detail.

\begin{figure}[htbp]
\centering
\includegraphics[scale=0.5]{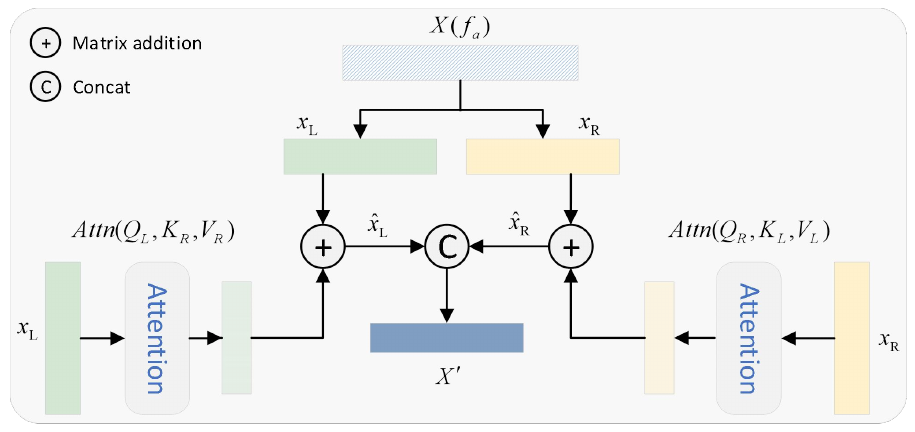}
\caption{Detailed structural diagram of the Stereo-Aware Attention Module.}
\label{figure2}
\end{figure}

\subsection{Feature Extraction}

Two separate CNN encoders are used for visual and audio modalities. The visual input is either an RGB image or a depth image of shape $128 \times 128 \times 3$ or $128 \times 128 \times 1$, respectively. Following the preprocessing strategy in~\cite{chen2020soundspaces}, the binaural audio input is transformed into spectrograms and stacked along the channel dimension: $65 \times 65 \times 2$ for Log-Mel or $65 \times 26 \times 2$ for STFT.

The visual and audio encoders share the same architecture, consisting of three convolutional layers (Conv8x8, Conv4x4, Conv3x3) followed by a fully-connected layer (Linear) with output dimension 512. A ReLU activation follows each convolutional layer. The resulting modality features are denoted as $f_v \in \mathbb{R}^d$ and $f_a \in \mathbb{R}^d$, where $d = 512$.

To enhance the spatial understanding of the audio input, we introduce the \textbf{Stereo-Aware Attention Module (SAM)} in the audio encoder (Fig.~\ref{figure2}). After the CNN layers, the feature map $X \in \mathbb{R}^{B \times C \times H \times W}$ ($C=64$) is split into left and right channel representations:

\[
x_L = X[:, :C/2, :, :], \quad x_R = X[:, C/2:, :, :]. \tag{1}
\]

A bidirectional cross-attention mechanism is applied:

\begin{align}
\hat{x}_L &= \text{Proj}\left(\text{Attn}(Q_L, K_R, V_R)\right) + x_L, \tag{2} \\
\hat{x}_R &= \text{Proj}\left(\text{Attn}(Q_R, K_L, V_L)\right) + x_R, \tag{3} \\
X' &= [\hat{x}_L; \hat{x}_R]. \tag{4}
\end{align}

Attn is the standard scaled dot-product attention, and Proj denotes a $1 \times 1$ convolution to restore the original channel size. This module enables explicit modeling of inter-channel spatial relationships, improving spatial awareness and sound source localization.

\begin{figure}[htbp]
\centering
\includegraphics[scale=0.5]{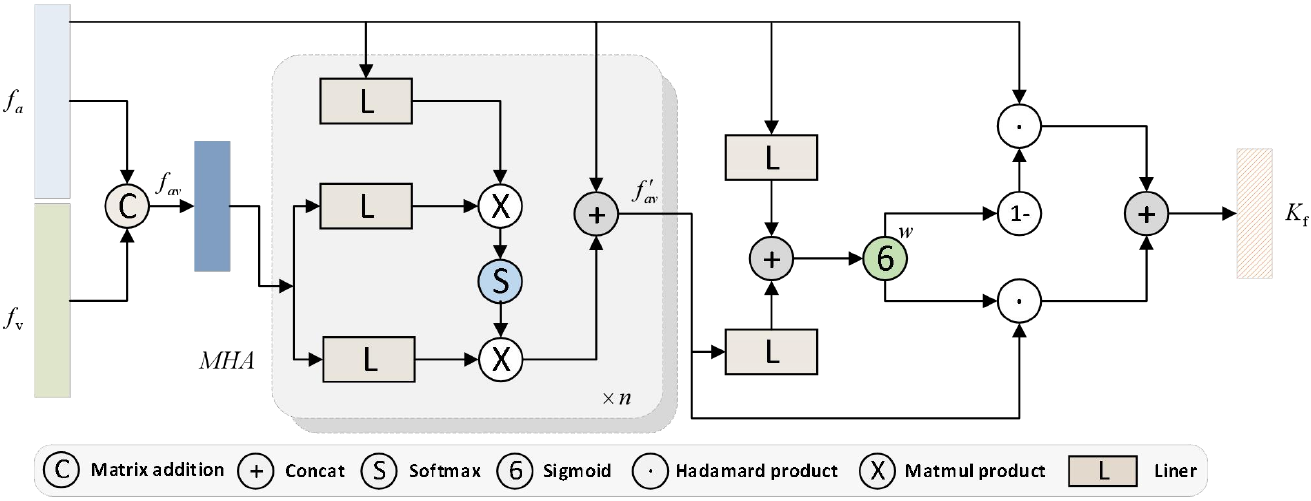}
\caption{Detailed structural diagram of the Audio-Guided Dynamic Fusion Module.}
\label{figure3}
\end{figure}

\subsection{Dynamic Feature Fusion}

As illustrated in Fig.~\ref{figure3}, we propose the \textbf{Audio-Guided Dynamic Fusion Module (AGDF)} to perform cross-modal attention-based fusion. The audio feature $f_a$ is first embedded into a 768-dimensional vector ($f_a$ is processed by SAM). The concatenated audio-visual feature $f_{av}$ serves as Key/Value, while $f_a$ is used as the Query in a multi-head attention (MHA) mechanism to generate an enhanced representation $f_{av}'$.

A learnable gating mechanism then dynamically balances $f_a$ and $f_{av}'$, producing a final fused representation $K_f$:

\begin{align}
f_{av} &= \text{Cat}(f_a, f_v), \tag{5} \\
f'_{av} &= \text{MHA}(f_a, f_{av}), \tag{6} \\
\omega &= \delta(f'_{av} W'_{av} + f_a W_a + b), \tag{7} \\
K_f &= \omega \odot f'_{av} + (1 - \omega) \odot f_a. \tag{8}
\end{align}

Here, $W'_{av}, W_a \in \mathbb{R}^{d_m \times 1}$ are learnable parameters, $b$ is the bias term, $\delta(\cdot)$ is a sigmoid activation, and $\odot$ denotes element-wise multiplication. $K_f$ is the final fused representation used for navigation policy input.

\subsection{Policy Learning}

To capture temporal dependencies, we use a GRU network to model the sequential dynamics of the fused features. At each step $t$, the GRU takes the current fused feature and previous hidden state $h_{t-1}$ to produce a new hidden state $h_t$ and observation embedding $o_t$. This is fed into an Actor-Critic policy network where: the \textbf{Actor} estimates the policy $\pi_\theta(a_t | o_t, h_{t-1})$; the \textbf{Critic} estimates the value function $V_\theta(o_t, h_{t-1})$.

The action space is discrete: \textit{Move Forward}, \textit{Turn Left}, \textit{Turn Right}, and \textit{Stop}. The agent selects the optimal action at each step to reach the goal efficiently. Training is performed using the Proximal Policy Optimization (PPO)~\cite{schulman2017proximal} algorithm, optimizing the following objectives: Clipped Policy Loss; Value Loss (weighted by 0.5); Entropy Bonus (weighted by 0.01).

We adopt Generalized Advantage Estimation (GAE) to reduce the variance of the temporal difference error while maintaining low bias~\cite{wang2023learning}. The reward design balances efficiency and goal success: (1) $+10$ reward for executing \textit{Stop} correctly at the target; (2) $-0.01$ time penalty per action; (3) $+1$ reward if the action reduces geodesic distance to the goal; (4) Negative reward if the distance increases. This reward shaping encourages efficient and successful navigation.

\section{Experiments and Analysis}

To validate our proposed method's effectiveness and generalization ability in the AVN task, we conduct systematic experiments on two realistic 3D simulation datasets and compare our approach with several mainstream baselines. This section presents the experimental setup, quantitative and qualitative results, and a visualized analysis of the agent's trajectories.

\subsection{Experimental Setup}

\subsubsection{Environment and Datasets.}
We evaluate our proposed method in the SoundSpaces~\cite{chen2020soundspaces} environment using the Habitat simulator~\cite{savva2019habitat} with two realistic 3D indoor datasets: Replica~\cite{replica19arxiv} and Matterport3D~\cite{Matterport3D}. Following the setting in~\cite{chen2020soundspaces}, the datasets are split into train/val/test sets. Replica contains 9/4/5 scenes, while sound classes split Matterport3D into 73/11/18 categories~\cite{wang2023learning}. Replica is relatively small with an average scene area of 47.24 $m^2$, whereas Matterport3D offers more complex and spacious environments with an average of 517.34 $m^2$~\cite{wang2023learning}.

Sound sources in SoundSpaces emit omnidirectional signals~\cite{chen2020soundspaces}, which are convolved with precomputed RIRs to generate binaural audio signals perceived from the agent’s current orientation. We consider two evaluation settings: \textit{heard sounds} and \textit{unheard sounds}. The \textit{heard} category includes both single and variable sound sources, while the \textit{unheard} setting—being more challenging—is our primary focus.

\begin{table}[ht]
\centering
\small
\caption{Performance comparison on Replica and Matterport3D datasets under two settings: heard (unseen scenes, seen sounds) and unheard (unseen scenes and sounds), using 78 training sound categories.Taking depth input as an example.}
\label{tab:table1}
\resizebox{0.95\textwidth}{!}{ 
\setlength{\tabcolsep}{4pt}
\begin{tabular}{c|ccc|ccc|ccc|ccc}
\toprule
\multirow{2}{*}{\textbf{Method}} & \multicolumn{6}{c|}{\textbf{Replica}} & \multicolumn{6}{c}{\textbf{Matterport3D}} \\
\cmidrule{2-13}
& \multicolumn{3}{c|}{Heard} & \multicolumn{3}{c|}{Unheard} & \multicolumn{3}{c|}{Heard} & \multicolumn{3}{c}{Unheard} \\
& SR$\uparrow$ & SPL$\uparrow$ & SNA$\uparrow$ & SR$\uparrow$ & SPL$\uparrow$ & SNA$\uparrow$ & SR$\uparrow$ & SPL$\uparrow$ & SNA$\uparrow$ & SR$\uparrow$ & SPL$\uparrow$ & SNA$\uparrow$ \\
\midrule
Random Agent~\cite{chen2023omnidirectional} & 18.5 & 4.9 & 1.8 & 18.5 & 4.9 & 1.8 & 9.1 & 2.1 & 0.8 & 9.1 & 2.1 & 0.8 \\
Direction Follower~\cite{chen2023omnidirectional} & 72.0 & 54.7 & 41.1 & 17.2 & 11.1 & 8.4 & 41.2 & 32.3 & 23.8 & 18.0 & 13.9 & 10.7 \\
Frontier Waypoints~\cite{chen2023omnidirectional} & 63.9 & 44.0 & 35.2 & 14.8 & 6.5 & 5.1 & 42.8 & 30.6 & 22.2 & 16.4 & 10.9 & 8.1 \\
SAVi~\cite{chen2021semantic} & 54.0 & 45.1 & 30.8 & 33.9 & 27.5 & 17.2 & 40.3 & 29.1 & 13.0 & 29.5 & 20.4 & 9.6 \\
AV-NaV~\cite{chen2020soundspaces} & 88.9 & 64.5 & 44.1 & 47.3 & 34.7 & 14.1 & 66.2 & 44.8 & 27.3 & 33.5 & 21.9 & 10.4 \\
AGSA (Ours) & \textbf{93.2} & \textbf{75.5} & \textbf{52.0} & \textbf{48.3} & \textbf{36.6} & \textbf{22.4} & \textbf{70.0} & \textbf{54.1} & \textbf{30.0} & \textbf{36.5} & \textbf{26.2} & \textbf{13.1} \\
\bottomrule
\end{tabular}
} 
\end{table}

\subsubsection{Evaluation Metrics.}

To comprehensively evaluate navigation performance, we adopt three mainstream metrics:~(1) \textbf{Success Rate (SR)}: The proportion of episodes in which the agent successfully navigates to the sound source and executes the \texttt{Stop} action within the step limit. (2) \textbf{Success weighted by Path Length (SPL)}: A metric that accounts for path efficiency, comparing the ratio between the shortest possible path and the actual path taken in successful episodes~\cite{yuan2024exploring}. (3) \textbf{Success weighted by Number of Actions (SNA)}: A metric that reflects action efficiency, encouraging the agent to complete the task using fewer movements and penalizing redundant behaviors such as in-place rotations~\cite{wang2023learning}.

\begin{table}[t]
\centering
\small
\caption{Evaluate each module's contribution to navigation under the audio-only (blind) setting. The Same setting uses only "telephone" sounds for training and testing, while the Multiple setting follows the earlier heard setting.}
\label{tab:table2}
\resizebox{1\textwidth}{!}{ 
\begin{tabular}{c|ccc|ccc|ccc|ccc}
\toprule
\multirow{3}{*}{\textbf{Method}} 
& \multicolumn{6}{c|}{\textbf{Replica}} 
& \multicolumn{6}{c}{\textbf{Matterport3D}} \\
\cmidrule{2-13}
& \multicolumn{3}{c|}{\textbf{Same}} & \multicolumn{3}{c|}{\textbf{Multiple}} 
& \multicolumn{3}{c|}{\textbf{Same}} & \multicolumn{3}{c}{\textbf{Multiple}} \\
\cmidrule{2-13}
& SR$\uparrow$ & SPL$\uparrow$ & SNA$\uparrow$ 
& SR$\uparrow$ & SPL$\uparrow$ & SNA$\uparrow$ 
& SR$\uparrow$ & SPL$\uparrow$ & SNA$\uparrow$ 
& SR$\uparrow$ & SPL$\uparrow$ & SNA$\uparrow$ \\
\midrule
AV-NaV    & 93.8 & 67.3 & 31.0 & 73.0 & 44.9 & 19.9 & 64.2 & 43.8 & 20.5 & 58.2 & 35.2 & 15.6 \\
Only SAM  & 96.4 & 70.8 & 34.4 & 78.1 & 51.7 & 21.5 & 65.5 & \textbf{45.6} & 19.5 & 59.8 & \textbf{40.1} & \textbf{18.0}   \\
Only AGDF & \textbf{96.9} & 72.0 & 35.6 & 71.7 & 47.3 & 19.5 & 65.0 & 44.3 & \textbf{21.1} & \textbf{61.3} & 40.1 & 17.7   \\
Ours       & 96.6 & \textbf{77.3} & \textbf{46.2} & \textbf{86.3} & \textbf{63.2} & \textbf{31.1} & \textbf{67.0} & 45.3 & 20.3 & 60.0 & 40.0 & 17.0 \\
\bottomrule
\end{tabular}
}
\end{table}

\subsubsection{Baselines.}

We compare our approach against the following representative baselines:

\begin{itemize}
    \item [$\bullet$] \textbf{Random}: Random action sampling at each time step from the action space, terminated only by the Stop action. This serves as a lower-bound baseline.
    \item [$\bullet$] \textbf{Direction Follower}: Estimates the Direction of Arrival (DoA) of the sound source and navigates towards an intermediate waypoint located $K$ meters in that direction~\cite{wang2023learning}.
    \item [$\bullet$] \textbf{Frontier Waypoints}: Uses the intersection of DoA and the frontier of explored regions to define intermediate goals~\cite{chaplot2020learning}.
    \item [$\bullet$] \textbf{SAVi}~\cite{chen2021semantic}: Originally designed for semantic audio-visual navigation with a transformer backbone. Since our task lacks explicit semantic targets and continuously emits sound during the episode, we adapt it by removing the goal descriptor network.
    \item [$\bullet$] \textbf{AV-NaV}~\cite{chen2020soundspaces}: An end-to-end reinforcement learning approach that navigates using audio and visual inputs without any maps or prior knowledge, representing a perception-driven strategy.
\end{itemize}

\subsection{Results and Analysis}

This section provides a detailed analysis of experimental results, including quantitative evaluation, ablation studies, and visualization of navigation trajectories.

\subsubsection{Quantitative Results.}

As shown in Table~\ref{tab:table1}, we systematically evaluate the navigation performance of the proposed AGSA method on the Replica and Matterport3D datasets under two evaluation settings: heard (unseen scenes, seen sounds) and unheard (unseen scenes and unseen sounds). 

Our method outperforms all baselines across all metrics (SR, SPL, SNA) in the challenging setting, demonstrating robust overall navigation performance. For example, on Replica, we achieve a 93.2\% SR, 75.5 SPL, and 52.0 SNA, significantly surpassing other methods. This indicates that our AGSA effectively leverages complementary audio-visual cues for improved target localization and efficient path planning.

Under the more challenging, unheard setting, our method performs significantly better than existing methods, for instance, on Matterport3D. Under the unheard setting, AGSA achieves 36.5 in SR and 26.2 in SPL, an improvement of approximately 3.0 and 4.3, respectively, over AV-NaV~\cite{chen2020soundspaces}. This suggests that AGSA exhibits strong generalization ability and can effectively handle the challenges posed by unseen sound categories.

We also highlight AGSA's outstanding performance under vision-free conditions (i.e. blind navigation), as shown in Table~\ref{tab:table2}. For instance, in the Replica-Same setting, the SR reaches 96.6\%, even in the more challenging Multiple setting, it still achieves 86.3\% in SR and a SPL of 63.2, significantly outperforming AV-NaV's SPL of 44.9, with a relative improvement of nearly 40\%. 
These substantial gains demonstrate that AGSA can maintain superior spatial awareness and auditory-based localization capabilities even without visual input, validating its robustness without complete modalities.

\subsubsection{Ablation Study.}

We perform ablation studies on the two key modules in our framework: the Stereo-Aware Attention Module (SAM) and the Audio-Guided Dynamic Fusion Module (AGDF). Results are shown in Table~\ref{tab:table3}.

Removing both SAM and AGDF significantly degrades performance. For example, in Replica-Multiple, SR drops from 93.2\% to 88.9\%, and SNA from 52.0 to 44.1, showing impaired understanding of audio signals and fusion capabilities. Analyzing their contributions, we find:~(1) Removing SAM (Ours w/o SAM) leads to a noticeable drop in SPL and SNA (e.g. SNA drops from 34.3 to 20.9 in Matterport3D-Same), highlighting SAM’s role in spatial cue extraction from binaural audio.~(2) Removing AGDF (Ours w/o AGDF) mainly affects SR (e.g. SR drops from 70.0\% to 66.3\% in Matterport3D-Multiple), indicating that AGDF facilitates robust decision-making via adaptive multi-modal fusion, especially when modalities conflict.

SAM and AGDF provide complementary advantages: SAM enhances spatial auditory perception, while AGDF ensures reliable fusion of audio-visual signals, jointly improving the robustness and generalization of the navigation strategy.

\begin{table}[t]
\centering
\small
\caption{Ablation study results on Replica and Matterport3D datasets under Same and Multiple sound source settings. Taking depth input as an example.}
\label{tab:table3}
\resizebox{1\textwidth}{!}{ 
\begin{tabular}{c|ccc|ccc|ccc|ccc}
\toprule
\multirow{3}{*}{\textbf{Ablation}} 
& \multicolumn{6}{c|}{\textbf{Replica}} 
& \multicolumn{6}{c}{\textbf{Matterport3D}} \\
\cmidrule{2-13}
& \multicolumn{3}{c|}{\textbf{Same}} & \multicolumn{3}{c|}{\textbf{Multiple}} 
& \multicolumn{3}{c|}{\textbf{Same}} & \multicolumn{3}{c}{\textbf{Multiple}} \\
\cmidrule{2-13}
& SR$\uparrow$ & SPL$\uparrow$ & SNA$\uparrow$ 
& SR$\uparrow$ & SPL$\uparrow$ & SNA$\uparrow$ 
& SR$\uparrow$ & SPL$\uparrow$ & SNA$\uparrow$ 
& SR$\uparrow$ & SPL$\uparrow$ & SNA$\uparrow$ \\
\midrule
Ours w/o SAM and AGDF & 90.1 & 75.6 & 45.3 & 88.9 & 64.5 & 44.1 & 70.3 & 55.2 & 32.6 & 66.2 & 44.8 & 27.3 \\
Ours w/o AGDF          & 91.7 & 79.9 & \textbf{53.5} & 91.2 & 73.2 & 45.5 & 71.2 & 56.8 & 33.0 & 66.3 & 53.2 & 27.2 \\
Ours w/o SAM         & 93.1 & 79.0 & 50.8 & 90.2 & 72.4 & 41.1 & 63.0 & 43.2 & 20.9 & 67.8 & 51.7 & 27.6 \\
Ours                  & \textbf{95.6} & \textbf{80.1} & 52.0 & \textbf{93.2} & \textbf{75.5} & \textbf{52.0} & \textbf{72.1} & \textbf{57.8} & \textbf{34.3}   & \textbf{70.0} & \textbf{54.1} & \textbf{30.0} \\
\bottomrule
\end{tabular}
}
\end{table}

\begin{figure}[htbp]
\centering
\includegraphics[scale=0.6]{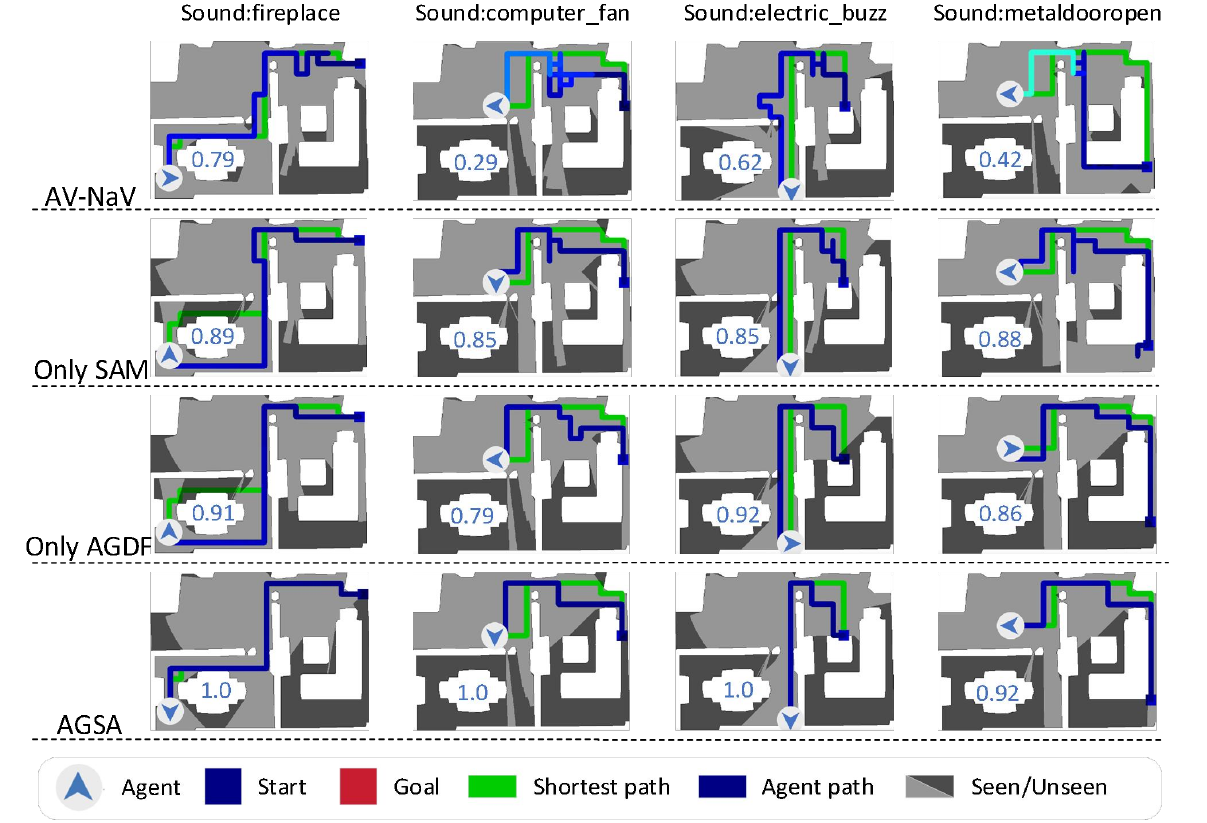}
\caption{Visualizing agent navigation trajectories under different sound settings. The numbers in each image represent the corresponding SPL values.}
\label{figure4}
\end{figure}

\subsubsection{Visualization Analysis.}

As illustrated in Figure~\ref{figure4}, we visualize representative navigation cases in the Replica dataset under various sound environments to compare agent trajectories across methods.

Our method consistently enables agents to quickly identify and move toward the target direction along nearly optimal paths. The resulting trajectories are smooth and efficient, with minimal deviation or in-place hesitation, indicating strong localization and planning capabilities. In contrast, baseline trajectories often show inefficient movements, detours, and frequent directional corrections, leading to longer episodes and reduced success rates. This highlights models' challenges, such as a lack of spatial auditory modeling or dynamic fusion mechanisms.

Overall, the trajectory visualizations demonstrate the practical impact of our SAM and AGDF modules: SAM equips the agent with precise spatial hearing, while AGDF adaptively balances modal contributions, resulting in a stable and effective navigation policy.

\section{Conclusion}

This study focuses on spatial perception and modality fusion in the Audio-Visual Navigation (AVN) task. It proposes an effective architecture, AGSA, which integrates a Stereo-Aware Attention Module (SAM) and an Audio-Guided Dynamic Fusion Module (AGDF). These modules help the agent more accurately determine the direction of the sound source and achieve adaptive fusion and dynamic adjustment between audio and visual modalities. 

We conducted systematic experiments in two realistic 3D simulation environments, Replica and Matterport3D, and achieved excellent performance under heard and unheard sound category settings. AGSA significantly outperforms existing methods in core metrics such as SR and SPL. Ablation studies further validate the effectiveness of each proposed module, showing that our method can substantially reduce path deviations and ineffective movements, leading to more efficient navigation strategies. 

In future work, we plan to explore improving model performance in more challenging AVN scenarios, such as multi-agent audio-visual navigation and navigation in more complex environments.

\section*{Acknowledgements}

This research was financially supported by the National Natural Science Foundation of China (Grants Nos. 62463029, 62472368, and 62303259) and the Natural Science Foundation of Tianjin (Grant No. 24JCQNJC00910).
%
\bibliographystyle{splncs04}
\bibliography{reference}

\end{document}